\def\year{2020}\relax
\documentclass[letterpaper]{article} 
\usepackage{aaai20}  
\usepackage{times}  
\usepackage{helvet} 
\usepackage{courier}  
\usepackage[hyphens]{url}  
\usepackage{graphicx} 

\usepackage{multirow}
\usepackage{amsfonts}
\usepackage[fleqn]{amsmath}
\usepackage[noend]{algpseudocode}
\usepackage{algorithmicx,algorithm}

\title{An Iterative Polishing Framework based on Quality Aware Masked Language Model for Chinese Poetry Generation }
\author{Liming Deng,\textsuperscript{\rm 1} Jie Wang,\textsuperscript{\rm 1} Hangming Liang,\textsuperscript{\rm 1} Hui Chen,\textsuperscript{\rm 1} \\ \Large \textbf{Zhiqiang Xie,\textsuperscript{\rm 3}\thanks{This work was done when Zhiqiang Xie was at Ping An Technology}  Bojin Zhuang,\textsuperscript{\rm 1} Shaojun Wang,\textsuperscript{\rm 1} Jing Xiao\textsuperscript{\rm 2}}\\
\textsuperscript{\rm 1}Ping An Technology\\ 
\textsuperscript{\rm 2}Ping An Insurance (Group) Company of China\\
\textsuperscript{\rm 3}University of Science and Technology of China\\
dengliming777@pingan.com.cn, photonicsjay@163.com
}
\begin{document}
\maketitle
\begin{abstract}
Owing to its unique literal and aesthetical characteristics, automatic generation of Chinese poetry is still challenging in Artificial Intelligence, which can hardly be straightforwardly realized by end-to-end methods. In this paper, we propose a novel iterative polishing framework for highly qualified Chinese poetry generation. In the first stage, an encoder-decoder structure is utilized to generate a poem draft. Afterwards, our proposed Quality-Aware Masked Language Model (\textbf{QA-MLM}) is employed to polish the draft towards higher quality in terms of linguistics and literalness.
Based on a multi-task learning scheme, \textbf{QA-MLM} is able to determine whether polishing is needed based on the poem draft. Furthermore, \textbf{QA-MLM} is able to localize improper characters of the poem draft and substitute with newly predicted ones accordingly.
 Benefited from the masked language model structure, \textbf{QA-MLM} incorporates global context information into the polishing process, which can obtain more appropriate polishing results than the unidirectional sequential decoding. Moreover, the iterative polishing process will be terminated automatically when \textbf{QA-MLM} regards the processed poem as a qualified one. Both human and automatic evaluation have been conducted, and the results demonstrate that our approach is effective to improve the performance of encoder-decoder structure.
\end{abstract}

\section{Introduction}
\label{sec:first}
\noindent Chinese Poetry, originated from people's production and life, has a long history.
The poetry is developed from few characters, vague rules to some fixed characters and lines with stable rules and forms.
The rules like tonal pattern, rhyme scheme lead to poems easy to be read and remembered.
The great poems, which touch millions of people at heart across the space and time, should unify the concise form, refined language and rich content together to guarantee the long-term prosperity. 
Writing great poems are not easy, which require strong desire for poets to express their feelings, views or thoughts and then to choose characters and build sentence carefully.

Poets are always regarded as genius with great talents and well trained in writing poems.
It is hard to write a poem for ordinary people, let alone to computers.
Although many works \cite{gervas2001expert,ghazvininejad2016generating,yi2018chineseA,li2018generating} have been conducted for automatic poetry generation and poetic rules and forms can be learned partially, the large gaps remain in the meaningfulness and coherence of generated poems.

In this paper, we focus on the automatic Chinese poetry generation and aim to fill these gaps.
We notice that poets would first write a poem draft and then polish the draft many times to a perfect one.
There is a popular story about polishing poem by Dao Jia, a famous poet in Tang Dynasty, who influences many later poets in polishing their poems intensively.
Motivated by the writing poem process of poets, we aim to imitate this process and improve the coherence and meaningfulness of primitive poems.
However, it is challenging for computer algorithms to automatically polish the poem draft to an excellent one.
The computer algorithms are unable to choose  the characters and sentences like poets with intuition and comprehensive understanding of the characters, which are only good at calculating the probability of characters and picking up ones with maximum probability from vocabulary.
There are three key issues to be addressed for the polishing framework.
\begin{itemize}
\item Whether the text need to be polished, and when should we stop the iterative polishing process?
\item Which characters in the text are improper and need to be replaced with better ones?
\item How to obtain the better ones?
\end{itemize}
To address these key issues and further improve the quality of generated poem, we propose a Quality-Aware Masked Language Model (\textbf{QA-MLM}) to implement an iterative polishing process.
To the best of our knowledge, this is the first work to solve the three key issues in polishing framework with one elegant model.

Our idea originates from the BERT \cite{devlin2018bert} with two-task learning schema, and we modify the tasks to aware of text quality and further obtain appropriate characters to replace the low quality characters in the text.
With these two tasks, we can polish the generated poem draft iteratively, and the polishing process will be terminated automatically.
The main contributions of this paper are summarized as follows:
\begin{itemize}
\item Our proposed model \textbf{QA-MLM}, a novel application of BERT for poem refinement, which can judge the quality of poem and polish the bad characters in the poem iteratively.
    The polish process will be terminated automatically until the polished poem is regarded as a qualified one by our model.
\item The \textbf{QA-MLM} model can obtain high quality characters by incorporating both left and right context information, and overcomes the weakness of the unidirectional decoding that only consider one side context information for the next character prediction.
\item A two-stage poem generation has been proposed to strengthen the meaningfulness and coherence of generated poems.
      On the first stage, the encoder-decoder structure has been utilized to generate the poem draft.
      Specifically, the pre-trained BERT and the transformer decoder has been utilized for poem draft generation.
      The poem draft is further polished by our proposed \textbf{QA-MLM} model on the second stage.
\end{itemize}

\section{Related Work}
\label{sec:second}
Automatic poetry generation has been investigated for decades.
The early work focus on improving the grammatical rules and complying with poetic forms with template-based methods
\cite{gervas2001expert,wu2009} and evolution algorithms \cite{manurung2004evolutionary,zhou2010genetic}.
The statistical machine translation methods \cite{he2012generating} and text summarization methods \cite{yan2013poet}
have also been utilized for generating more natural and flexible poems.

As neural network demonstrates powerful ability for natural language representation \cite{bengio2003neural,goldberg2017neural},
different neural network structures have been utilized for poem generation and shown great advances.
The main structures are developed from vanilla recurrent neural network \cite{zhang2014chinese} to
bi-directional long short term memory network
 and bi-directional gated recurrent unit network \cite{wang2016chineseB,yi2017generating}.
The poem generation is widely interpreted as a sequence-to-sequence problem, which utilize the encoder-decoder framework
to encode the previous sequence and generate the later sequence with the decoder \cite{wang2016can,yi2017generating}.
To strengthen the relation between the encoder and decoder, the attention mechanism has been incorporated for poem generation
\cite{wang2016can,zhang2017flexible,yi2017generating}.
Besides, some tricks like working memory mechanism and salient-clue mechanism \cite{yi2018chineseA,yi2018chineseB}
have been proposed to improve the coherence in meanings and topics for poem generation.
Recently, the conditional variational autoencoder (C-VAE) has been utilized to generate the poem
\cite{yang2017generating,li2018generating}.
The C-VAE can obtain high quality poem with topic coherence to some extent.
Some bad cases are also reported by \cite{li2018generating}.

All the previous methods are generating the poem directly without any refinement.
The most similar work to our paper is \emph{i,poet} \cite{yan2016poet}, which has implemented an polishing framework
via encoding the writing intents repetitively to rewrite the poem.
This work empirically polishes all the characters that generated at previous iterative step, and assumes that the further encoding of the
writing intents would enhance the theme consistency.
Another similar work is the \emph{Deliberation Network} \cite{xia2017deliberation}, which has utilized a second decoder to generate the final sequence with the additional input of sequence that generated by the first decoder.
Followed by the \emph{Deliberation Network}, a more recent work \cite{zhang2019pretraining} has employed transformer decoder to implement the two-decoder polishing process for text summarization.
All these methods fail to sense the text quality and regard rewriting the whole sequence as polishing process, which are inefficiency and may replace qualified characters with low-quality ones.
Besides, the polishing process in these methods are heavily coupled with the initial draft generation process, which refers to not only the generated drafts but also the additional information that has been utilized in the draft generation.
Therefore, these polishing process cannot be utilized to polish the text drafts that generated separately by other models.

By contrast, our proposed \textbf{QA-MLM} model is different to the previous works significantly and implements the iteratively polishing process like humans.
Our model can first aware of the text quality and decide whether the text need to be polished.
Furthermore, our model can aware of the low-quality characters and pick them up to be polished with both left and right context information. Since our model can sense the text quality, the iterative polish step will be terminated automatically once the polished text has been regarded as qualified.
Our model only polishes a small part of low-quality characters instead of rewriting the whole text.
The polishing process implemented by our model is independent to the draft generation process, which can polish the draft generated by other separate model.
In this paper, we apply our proposed \textbf{QA-MLM} model for Chinese poetry polishing, which can significantly improve the quality of poem particular in terms of meaningfulness and coherence.
We will introduce our approach in the following sections.
\begin{figure}[ht]
\centering
      \includegraphics[width=0.98\columnwidth]{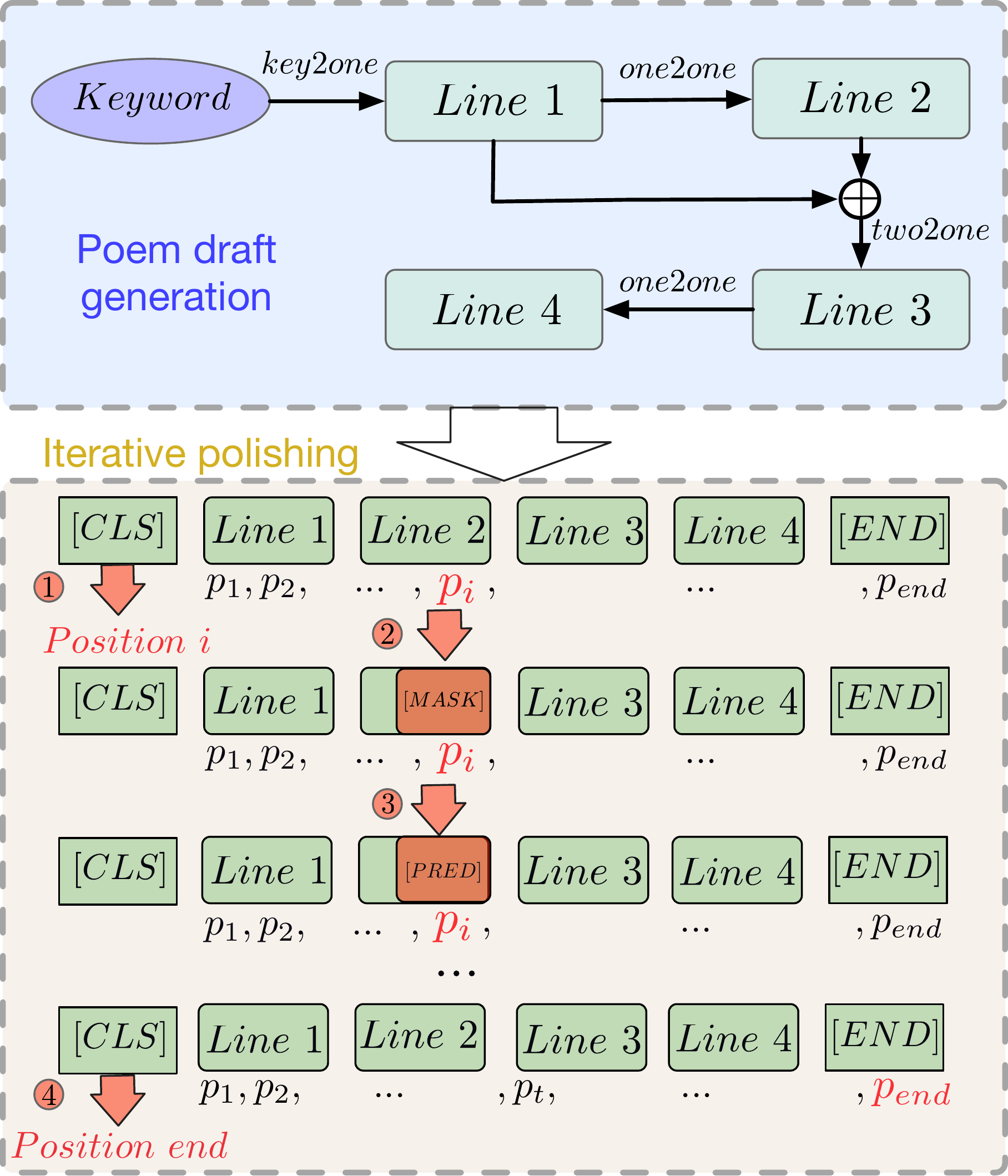}
      \caption{An overview of our poem generation approach, including poem draft generation and iterative polishing process.
      \textcircled{1}~Predict the character position with worst quality, \textcircled{2}~ Mask the worst quality character, \textcircled{3}~Predict the masked character and update it into the text accordingly, \textcircled{4}~Repeat the previous steps until no character needs to be polished.}
      \label{fig:ovv}
 \end{figure}
\section{Model Design}
\label{sec:third}
\subsection{Overview}
\label{ssec:3first}
We focus on the generation and polishing of quatrain, which is a famous kind of Chinese classic poetry with strict constraints of poetic rules and forms.
In general, the quatrain is with four poem lines, the number of characters for each line is equal, which is either five or seven.
The tone level and rhythm  are constrained with specific patterns \cite{wang2002summary}.
We follow the text-to-text generation paradigm \cite{wang2016chineseB,li2018generating} and generate the poem line by line.
The first poem line is generated by keywords or topic words, and the following lines are generated by the preceding lines or their combinations.
The key task turns into designing a proper model to generate the following lines with given keywords or preceding lines.

Inspired by the real procedure of writing poems for poets, we generate the poem lines with two stages.
The poem draft lines are first generated with encoder-decoder framework and then polished with our proposed \textbf{QA-MLM}.
The overall structure of our approach can be shown in Figure \ref{fig:ovv}.

\subsection{Input Representation}
\label{ssec:3second}
\begin{figure}[ht]
\centering
	  \includegraphics[width=0.94\columnwidth]{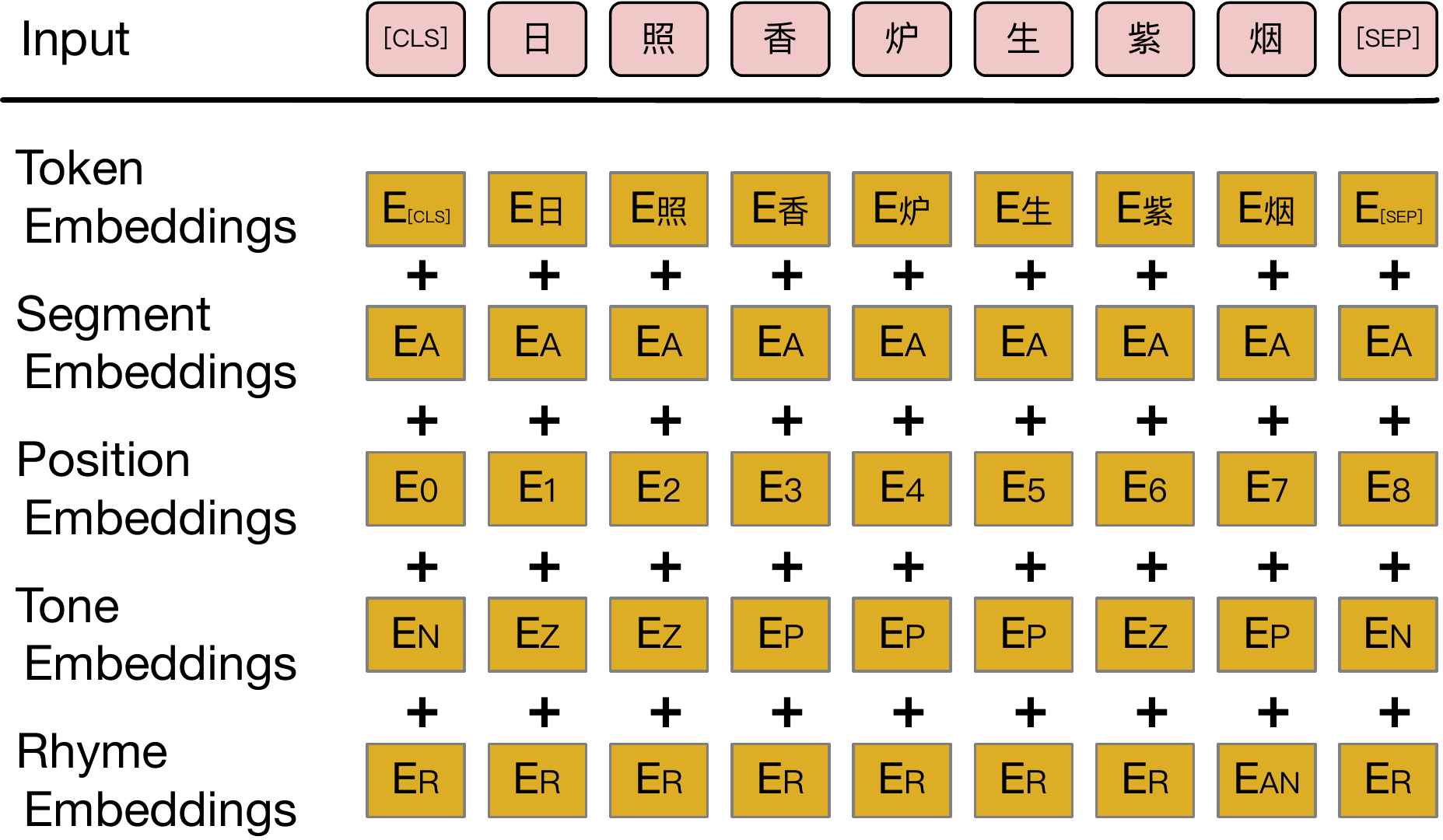}
      \caption{The input representation. The input embeddings is the sum of the token embeddings, segmentation embeddings, position embeddings and tone embeddings as well as rhyme embeddings. The $E_{N}$ and $E_{R}$ represent the token without tone or no need to embed the rhyme respectively.}
      \label{fig:embeddings}
 \end{figure}

Our input representation is similar to the embedding methods in BERT \cite{devlin2018bert}.
In addition to sum the token, segment and position embeddings as the representation, we also add the tonal
and rhythm embeddings into the input representation.
The tone of each character is either Ping (the level tone) or Ze (the downward tone) \cite{wang2002summary}.
We encode the tone with three categories due to some tokens like comma without any tone.
The rhythm of last character for each poem line will be encoded and we utilize the Thirteen Rhyme. \footnote{Classify the final vowels into thirteen categories according to the rhyme table.}
With the tone and rhythm are embedded into the representation, we can improve the poeticness of generated poem significantly without sacrifice much of the poem quality.
The visualization of our input representation is given in Figure \ref{fig:embeddings}.

 \begin{figure*}
  \centering
  \includegraphics[width=1.6\columnwidth]{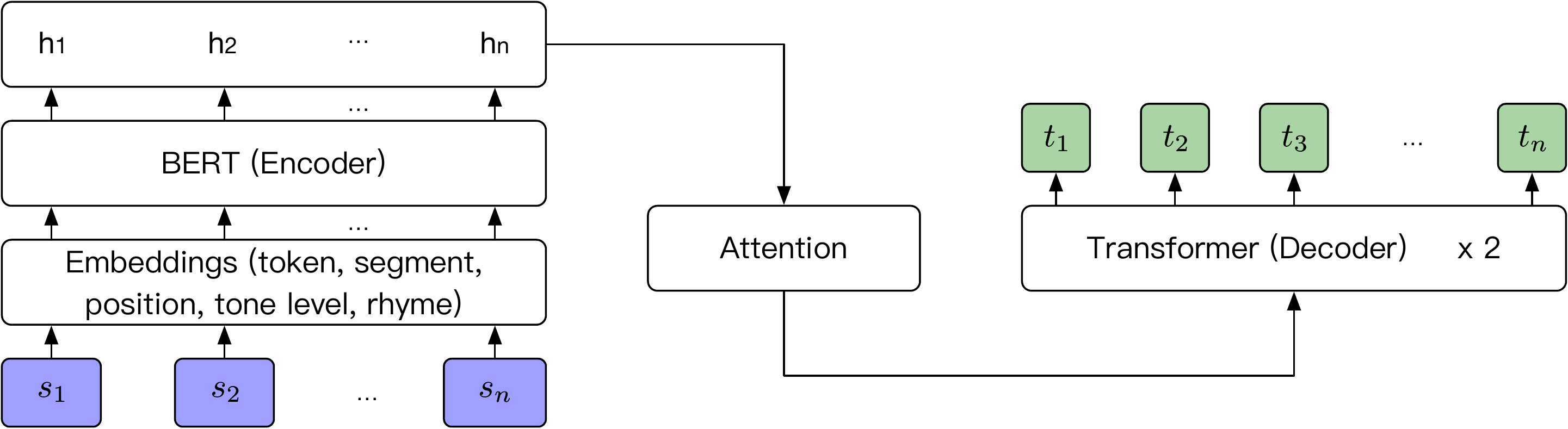}\\
  \caption{The sequence-to-sequence generation with BERT and Transformer decoder.}\label{fig:bt}
\end{figure*}
\subsection{Poem Draft Generation}
The poem draft can be generated via encoder-decoder structure.
This structure learns the relation between the target sequence $\mathbf{t}=\{t_1,t_2,...,t_n\}$ and
the source sequence $\mathbf{s}=\{s_1,s_2,...,s_m\}$.
The generation probability $P(\mathbf{t}|\mathbf{s};\Theta)$ can be obtained by Equation \eqref{eqa:pts},
where the $\Theta$ is model parameters and learned from the sequence pairs $(\mathbf{s},\mathbf{t}) \in (\mathbb{S},\mathbb{T})$.
We maximize the objective function in Equation \eqref{eqa:obj1}.
\begin{eqnarray}
    P(\mathbf{\hat{t}}|\mathbf{s};\Theta) &=& \prod_{j=1}^{n}P(\hat{t}_j|t_{<j},\mathbf{s};\Theta) \label{eqa:pts}\\
    L_{character} &=& \sum_{(\mathbf{s},\mathbf{t}) \in (\mathbb{S},\mathbb{T})} \log P(\mathbf{\hat{t}}|\mathbf{s};\Theta) \label{eqa:obj1}
\end{eqnarray}

As shown in Figure \ref{fig:bt}, the BERT is utilized as the encoder to represent the source sequence $\mathbf{s}$ with
vector $\mathbf{h}$, and the representation vector $\mathbf{h}$ is then fed to a two-layer transformer decoder
 to generate the target sequence $\mathbf{\hat{t}}$ \cite{devlin2018bert,zhang2019pretraining}.
The source sequence can be a keyword or poem lines, and the target sequence is the poem line that we want to generate.
All or part of previous sequence have been utilized as source sequence by previous works \cite{wang2016chineseB,yi2017generating}.
After carefully considering the relevance among poem lines and without making the generation system complicated,
we build three different models with the same structure for each poem line generation, namely: \textbf{key2one}, \textbf{one2one}, and \textbf{two2one}.

The \textbf{key2one} model is utilized to generate the first poem line with the input of keyword.
The \textbf{one2one} model is employed to generate the second poem line and the fourth poem line due to the similar relevance with their preceding poem lines.
As for generating the third poem line, we consider both the first and second poem lines via the \textbf{two2one} model.
The whole poem draft generation process can be visualized in the upper part of Figure \ref{fig:ovv}.

\subsection{Iterative Polishing}
\label{ssec:3fouth}
There is an obvious deficiency for the aforementioned encoder-decoder method \cite{xia2017deliberation}.
During the decoding, the character generated sequentially is affected by the previous characters and ignores the influence
of subsequent characters, as demonstrated in Equation \eqref{eqa:pts}.
Therefore, an iterative polishing framework which can utilize both previous and subsequent context information to polish center character is critical to obtain a semantic consistency and coherence poem.

We propose a quality aware masked language model (\textbf{QA-MLM}) to implement the iterative polishing process.
The quality aware reflects the model capability of distinguishing the character quality and deciding which character need to be polished.
Besides, our model can decide whether the text need to be polished and when should we stop the iterative polishing process.
The masked language model is to mask the low quality character first and then predict another character by referring the two-side context information.
The predicted character is assumed to be with better semantic consistency and coherence due to the consideration of both the previous and subsequent information.
Inspired by \cite{devlin2018bert}, we train the \textbf{QA-MLM} with two prediction tasks and apply it to polish the generated poem draft, as described in the following subsections.
\begin{figure}[ht]
  \centering
  \includegraphics[width=0.7\columnwidth]{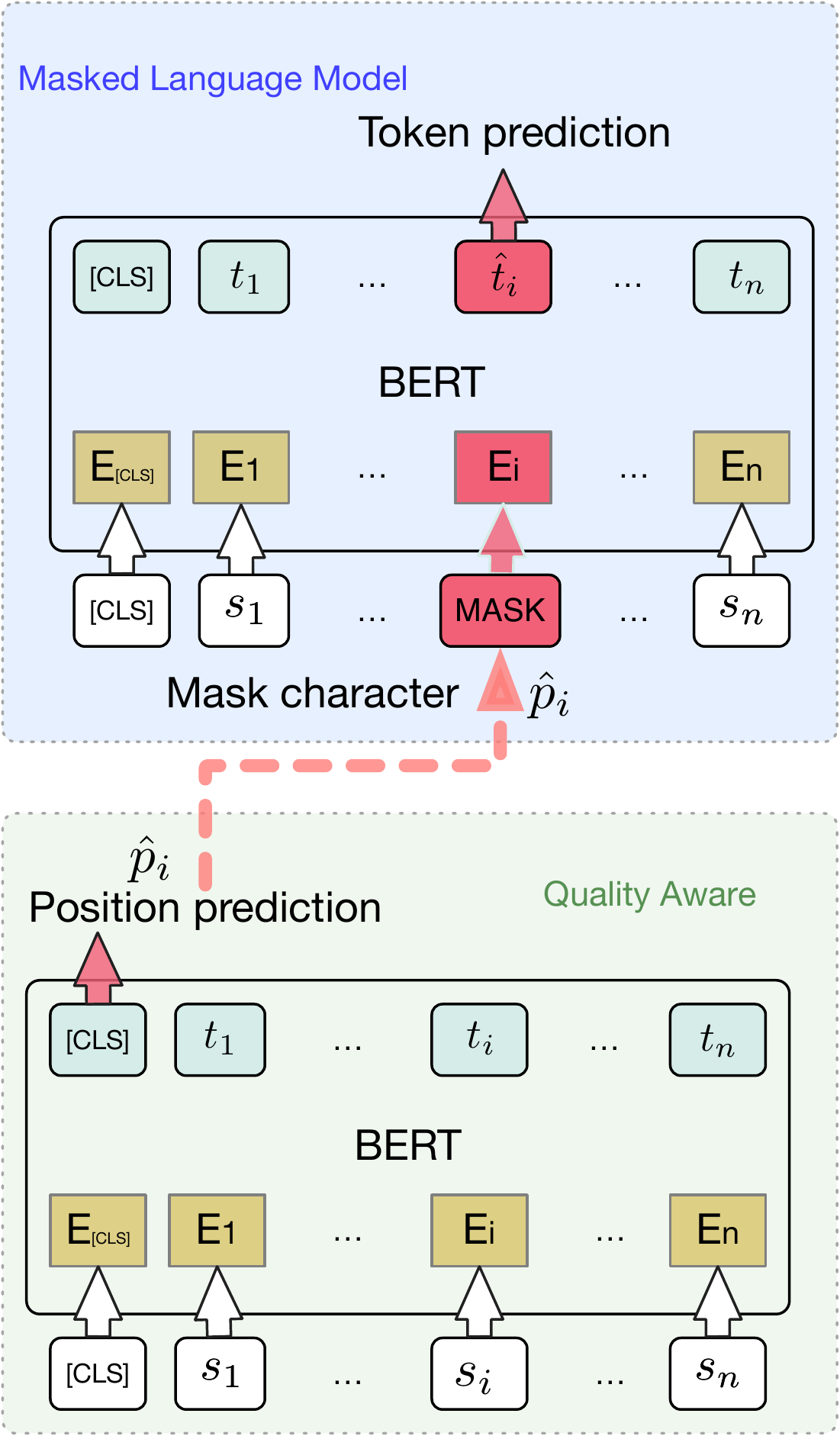}\\
  \caption{The structure of quality aware masked language model.}\label{fig:itpolish}
\end{figure}

\subsubsection{Prediction Tasks}

In order to provide reasonable solutions to the aforementioned three key issues in polishing framework, we design a quality prediction task and a masked language model task based on BERT \cite{devlin2018bert}.
Unlike the BERT, which learns from multi-task for context representation,
our proposed \textbf{QA-MLM} predicts the positions of low quality characters and then replaces the low quality characters with newly predicted ones for text refinement.
The structure of \textbf{QA-MLM} can be visualized in Figure \ref{fig:itpolish}.
The quality prediction task is to predict the character positions that the characters are with low quality.
We regard the original poem lines from poetry corpus are the gold standards, and any changing of the original poem lines would hurt the quality of the poem.
Therefore, we randomly replace the characters in original poem line with random tokens.
We denote $s_g$ as the original poem line, and the $s_c$ as the changed poem line.
The positions that have been replaced can be denoted as $p = [p_{i1},p_{i2},...,p_{ir}]$, where $ir<n$, and the real characters that have been replaced are $s_i=[s_{i1},s_{i2},...,s_{ir}]$.
The number of replaced positions $r$ reflects the learning capacity of the \textbf{QA-MLM} that how many mismatched characters can be learned.
A larger $r$ allows a more powerful model for polishing the bad poem lines intensively.
 However, the large $r$ would lead to the changed poem lines very bad and increases the training difficult.
It should be careful to choose an appropriate $r$ considering both the model capacity and learning quality.

In this work, each poem line is randomly replaced \footnote{Refers to both the positions and the tokens are randomly selected} according to the following rules:
\begin{itemize}
\item 60\% of the time: replace one character with random token,
eg., the original poem line $\mathbf{s_g}=[s_1,s_2,s_3,s_4,s_5,s_6,s_7]$ ($n=7$, for example) is changed to $\mathbf{s_c}=[s_1,s_2,\mathbf{s_{i1}},s_4,s_5,s_6,s_7]$ and the position label is $p = 3$, then the masked poem line is $\mathbf{s_m}=[s_1,s_2,\mathbf{[MASK]},s_4,s_5,s_6,s_7]$.

\item 20\% of the time: replace two characters with random tokens,
eg., the original poem line $\mathbf{s_g}=[s_1,s_2,s_3,s_4,s_5,s_6,s_7]$ is changed to $\mathbf{s_c}=[s_1,\mathbf{s_{i1}},s_3,s_4,s_5,\mathbf{s_{i2}},s_7]$ and the position label is $p = [2,6]$, then the masked poem line is $\mathbf{s_m}=[s_1,\mathbf{[MASK]},s_3,s_4,s_5,\mathbf{[MASK]},s_7]$.

\item 20\% of the time: keep the poem line unchanged, in this situation we set the position label as 0, namely $\mathbf{s_g}=\mathbf{s_c}$ and $p=0$, there is no need to mask the poem line $\mathbf{s_c}$ and no need to polish poem line $\mathbf{s_c}$, which infers $p_{end}=0$.
\end{itemize}

\subsubsection{Learning and Inference}
We can jointly learning the aforementioned two tasks to train \textbf{QA-MLM} by minimizing the following loss functions:
\begin{align}
    Loss_{q} = -&\sum_{\mathbf{s_c} \in \mathbb{S}_c}\sum_{j=i1}^{ir} \log   P(\hat{p}_j = p_j|\mathbf{s_c};\Theta) \label{eqa:obj3}\\
    Loss_{m} = -&\sum_{(\mathbf{s_m},\mathbf{s_g}) \in (\mathbb{S}_m,\mathbb{S}_g)}\sum_{j = i1}^{ir} \nonumber\\
    &\log P(\hat{s}_{m,j} = s_{g,j}|s_{m,\neq{j}};\Theta) \label{eqa:obj22}\\
    Loss_{total} =& Loss_{m} + \lambda Loss_{q} 
\end{align}
After learning our proposed \textbf{QA-MLM} over the constructed poem corpus $(\mathbb{S}_g,\mathbb{S}_c,\mathbb{S}_m)$, we can utilize the \textbf{QA-MLM} to polish the poem draft.
At the beginning of polishing process, the \textbf{QA-MLM} predicts the character position that with worst character quality.
If the predicted position $p_i$ is equal to $p_{end}$ (in our setting $p_{end}=0$), which means that all characters in the draft are qualified enough and no more any polishing, otherwise the character identified with the worst quality will be masked in the draft, and the masked draft will be further utilized for a better character prediction via \textbf{QA-MLM} model. %
The predicted character is regarded as more appropriate than the masked character due to the incorporation of two-side context information during the prediction.
Thus, we replace the masked character in sequence draft with the predicted character, and one polishing step is finished.
By repeating the above polishing step, the sequence draft can be iteratively polished many times until the end indicator $p_{end}$ is predicted.
At this time, the iterative polishing process will be terminated automatically, and the sequence draft has been polished completely.
The iterative polishing process can be shown in Algorithm \ref{alg:polish}.
\begin{algorithm}[ht]
\caption{\textbf{:} Iterative Polishing with \textbf{QA-MLM}} \label{alg:polish}
\begin{algorithmic}[1]
\State Perform iterative polishing on sequence draft $\mathbf{s_d}$ = [$s_{1}$,$s_{2}$,...,$s_{i}$,...,$s_{n}$]
\State Predict bad character position $p_{i}=\mathbf{QA}(\mathbf{s_d})$
\While{\textit{$p_{i} \ne p_{end} $  }}
\State $s_{i}$ $\leftarrow$ {\scriptsize [MASK]} , $\mathbf{s_d}$ = [$s_{1}$,$s_{2}$,...,{\scriptsize [MASK]},...,$s_{n}$]
\State Predict the new character $\hat{s}_{i}= \mathbf{MLM}(\mathbf{s_d})$
\State $\mathbf{s_d}$ $\leftarrow$ [$s_{1}$,$s_{2}$,...,$\hat{s}_{i}$,...,$s_{n}$]
\State $p_i$ $\leftarrow$  $\mathbf{QA}(\mathbf{s_d})$
\EndWhile
\State \Return  polished  $\mathbf{s_d}$
\end{algorithmic}
\end{algorithm}

The sequence draft can be a poem line or several poem lines and even a whole poem, our approach is capable of polishing all of them.
In this work, we polish the whole poem together, which incorporates the whole context information for inappropriate character prediction, and the inappropriate characters will be replaced with highly qualified characters to obtain semantic consistency and coherence.
\section{Experiments and Evaluations}
\begin{table*}
\begin{center}
\caption{\label{tab:bleu}The BLEU score results on different generated poem line with same extracted keyword or previous poem lines. \textbf{BL-1} and \textbf{BL-2} are the BLEU scores on unigrams and bigrams. }
	\begin{tabular}{|c||c c|c c|c c|c c|c c|}
		\hline
		\multirow{2}{*}{\textbf{Model}} &
		\multicolumn{2}{|c|}{\textbf{key $\rightarrow$ 1}} & \multicolumn{2}{|c|}{\textbf{1 $\rightarrow$ 2}} &
		\multicolumn{2}{|c|}{\textbf{1\&2 $\rightarrow$ 3}} & \multicolumn{2}{|c|}{\textbf{3 $\rightarrow$ 4}} &
		\multicolumn{2}{|c|}{\textbf{Average}}\\
		&
		\textbf{BL-1} & \textbf{BL-2} & \textbf{BL-1} & \textbf{BL-2} & \textbf{BL-1} &
		\textbf{BL-2} & \textbf{BL-1} & \textbf{BL-2} & \textbf{BL-1} & \textbf{BL-2}\\
		\hline
		\hline
		\textbf{AS2S} &
		{0.072} & {0.047} & {0.016} & {0.005} & {0.019} &
		{0.006} & {0.021} & {0.007} & {0.032} & {0.016}\\
		\textbf{AS2S-P} &
		{0.074} & {0.047} & {0.026} & {0.009} & {0.030} &
		{0.010}&{0.036}&{0.012}& \textbf{0.042} & \textbf{0.020}\\
		\textbf{CVAE-D} &
		{0.109} & {0.059} & {0.013} & {0.005} & {0.015} &
		{0.005}&{0.019}&{0.006}&{0.039}&{0.019}\\
		\textbf{CVAE-D-P} &
		{0.105} & {0.057} & {0.015} & {0.005} & {0.016} &
		{0.006}&{0.021}&{0.007}&{0.039}&{0.019}\\
		\textbf{B\&T} &
		{0.102} & {0.050} & 0.028 & 0.010 & 0.036 &
		{0.014}&{0.035}&{0.013}&\textbf{0.050}&\textbf{0.022}\\
		\textbf{B\&T-P} &
		{0.100} & {0.050} & {0.028} & {0.009} & {0.034} &
		{0.013}&{0.033}&{0.012}&{0.049}&{0.021}\\
		\hline
	\end{tabular}
\end{center}
\end{table*}
\subsection{Experimental Setup}
In this paper, we concentrate on the generation of Chinese quatrain with seven fixed characters for each line.
Our poetry corpus is consist of poems from \emph{Tang Dynasty}, \emph{Song Dynasty}, \emph{Yuan Dynasty}, \emph{Ming Dynasty} and \emph{Qing Dynasty}.
About 130,525 poems with total 905,790 number of poem lines are filtered from the poetry corpus.
Each filtered poem contain four or multiple of four poem lines, and each poem line with seven characters.
These poems are randomly split into three part for model training (90\%), validation (5\%) and testing (5\%).
Three models (\textbf{key2one}, \textbf{one2one}, and \textbf{two2one}) trained by different sequence-to-sequence pairs are utilized to generate the poem draft lines.

The BERT is selected as the encoder with 12 layers and initialized with the parameters pre-trained by \cite{devlin2018bert}.
The 2-layer transformer decoder is selected for the poem generation.
After the poem draft has been generated, the \textbf{QA-MLM} is proposed for the polishing.
The aware of poem quality is implemented by the quality task to predict the position character that with worst semantic quality.
In addition to the current total 28 positions for the seven-character quatrain, an end position $p_{end} = 0$ is added to indicate the end of iterative polishing.
The character located by the quality prediction task will be masked and then replaced with newly predicted one by masked language model task. 
The quality prediction task and the masked language model task are based on the 12-layer BERT and learned jointly.

The conventional RNN encoder-decoder structure with attention mechanism (\textbf{AS2S}) \cite{wang2016chineseB} and a more recent work \textbf{CVAE-D} \cite{li2018generating} are selected as the baselines for poem draft generation.
Besides, we also implement a more powerful encoder-decoder structure with pre-trained BERT and transformer decoder (\textbf{B\&T}) for poem draft generation.
The poem drafts are generated with the input of keywords or writing intents, and we follow the keywords extraction method adopted by \textbf{AS2S} \cite{wang2016chineseB}, which cuts the poem lines into several word segmentations by \emph{Jieba} \cite{sun2012jieba} and then utilizes the \emph{TextRank} \cite{mihalcea2004textrank} method to select keyword with the highest score.
The poems generated by the aforementioned three models are further polished with the proposed \textbf{QA-MLM}.
Both the automatical evaluation criteria and human evaluations have been conducted.
The following subsection will introduce the detail about the evaluations.
\begin{table*}[t!]
\centering
\caption{\label{tab:res}The evaluation results. \textbf{Sim12} refers to the similarity between first poem line and second poem line; \textbf{Sim34} refers to the similarity between the third poem line and the fourth poem line; \textbf{Sim2L} refers to the similarity between first two poem lines and the last two poem lines; \textbf{TA.} and \textbf{RA.} are the tone level predicted accuracy and the rhythm predicted accuracy respectively; \textbf{Con.}, \textbf{Flu.}, \textbf{Mea.}, and \textbf{Poe.} represent the \emph{Consistency}, \emph{Fluency}, \emph{Meaningfulness} and \emph{Poeticness} respectively.}%
	\begin{tabular}{|c||c c c|c c||c c c c|}
		\hline
		\multirow{2}{*}{\textbf{Model}} &
		\multicolumn{5}{|c||}{\textbf{Automatic Evaluation}} &
		\multicolumn{4}{|c|}{\textbf{Human Evaluation}}\\
		\cline{2-10}
		&
		\textbf{Sim12} & \textbf{Sim34} & \textbf{Sim2L} & \textbf{TA.} & \textbf{RA.} &
		\textbf{Con.} & \textbf{Flu.} & \textbf{Mea.} & \textbf{Poe.}\\
		\hline
		\hline
		\textbf{AS2S} &
		{0.479} & {0.487} & {0.648} & \textbf{0.539} & {0.121} &
		{2.46} & {2.37} & {2.35} & {2.28}\\
		\textbf{AS2S-P} &
		\textbf{0.484} & \textbf{0.495} & \textbf{0.650} & {0.517} & \textbf{0.124} &
		\textbf{2.64}&\textbf{2.63}&\textbf{2.59}&\textbf{2.59}\\
		\textbf{CVAE-D} &
		{0.494} & {0.500} & {0.651} & {0.521} & {0.086} &
		{2.62}&{2.50}&{2.55}&{2.42}\\
		\textbf{CVAE-D-P} &
		\textbf{0.499} & \textbf{0.507} & \textbf{0.653} & \textbf{0.524} & \textbf{0.091} &
		\textbf{2.75}&\textbf{2.72}&\textbf{2.74}&\textbf{2.64}\\
		\textbf{B\&T} &
		{0.500} & {0.516} & {0.640} & \textbf{0.976} & \textbf{0.956} &
		{3.01}&{2.99}&{3.06}&{2.88}\\
		\textbf{B\&T-P} &
		\textbf{0.502} & \textbf{0.519} & \textbf{0.642} & {0.962} & {0.841} &
		\textbf{3.14}&\textbf{3.19}&\textbf{3.24}&\textbf{3.09}\\
		\hline
	\end{tabular}
\end{table*}
\subsection{Evaluation Metrics}
It is difficult for computer to estimate the quality of poem.
Therefore, we utilize both automatic evaluation metrics and human judgements for model comparisons.
The automatic evaluation metrics including BLEU \cite{Papineni2002BLEU}, Similarity \cite{wieting2015towards}, tone accuracy and rhythm accuracy are adopted in this paper.

The \textbf{BLEU} is designed for machine translation and also widely adopted by many previous works \cite{zhang2014chinese,li2018generating} in poem generation.
The BLEU is to measure the overlapping of characters between the generated sentence and the referred sentence.
Unlike the machine translation, the generated sentence can be significantly different from the referred sentence but also regarded as high quality by human judgements.
The poem generation is more related to creativity and the generated poem may far away from the referred poem, which may lead to the comparison of BLEU score is trivial.
Therefore, we compare the BLEU score on one sentence instead of the whole poem.
Each sentence is generated by different approaches with the same keyword or sentence input.

The \textbf{Similarity} is aimed to automatically measure the coherency or consistency among poem lines.
The embedding of characters can partially reflect the similarities and we accumulate the embeddings of all the characters for each poem line for sentence-level embeddings \cite{wieting2015towards}.
Then, the similarity between two poem lines can be calculated by the $cosine$ function on the sentence-level embeddings.

The \textbf{Tone Accuracy} and \textbf{Rhythm Accuracy} are also employed for the evaluation.
The tone accuracy is the percentage that the tone level (Ping or Ze) is predicted correct to all the generated samples, and the rhythm accuracy is similar about the last character of each poem line that the rhyme is predicted correct.

The \textbf{Human Evaluation} is inevitable for poem evaluation, which is more reliable and credible than the automatic evaluation metrics.
Twenty well educated annotators are invited to evaluate the generated poems in four dimensions, namely \emph{Consistency},
\emph{Fluency}, \emph{Meaningfulness} and \emph{Poeticness} \cite{zhang2014chinese,li2018generating}.
Each dimension is rated using the 1 to 5 scale to represent from bad quality to excellent.
Each model generates one thousand poems and the poems are divided equally into twenty pieces.
To reduce the individual scoring bias, the poems rated by each participant are from all models, but the participant has no information about the model that each poem belongs to.
Therefore, we can obtain 6000 ($20\times 6 \times 50$) poem ratings.
\begin{figure*}[ht]
\centering
      \includegraphics[width=1.4\columnwidth]{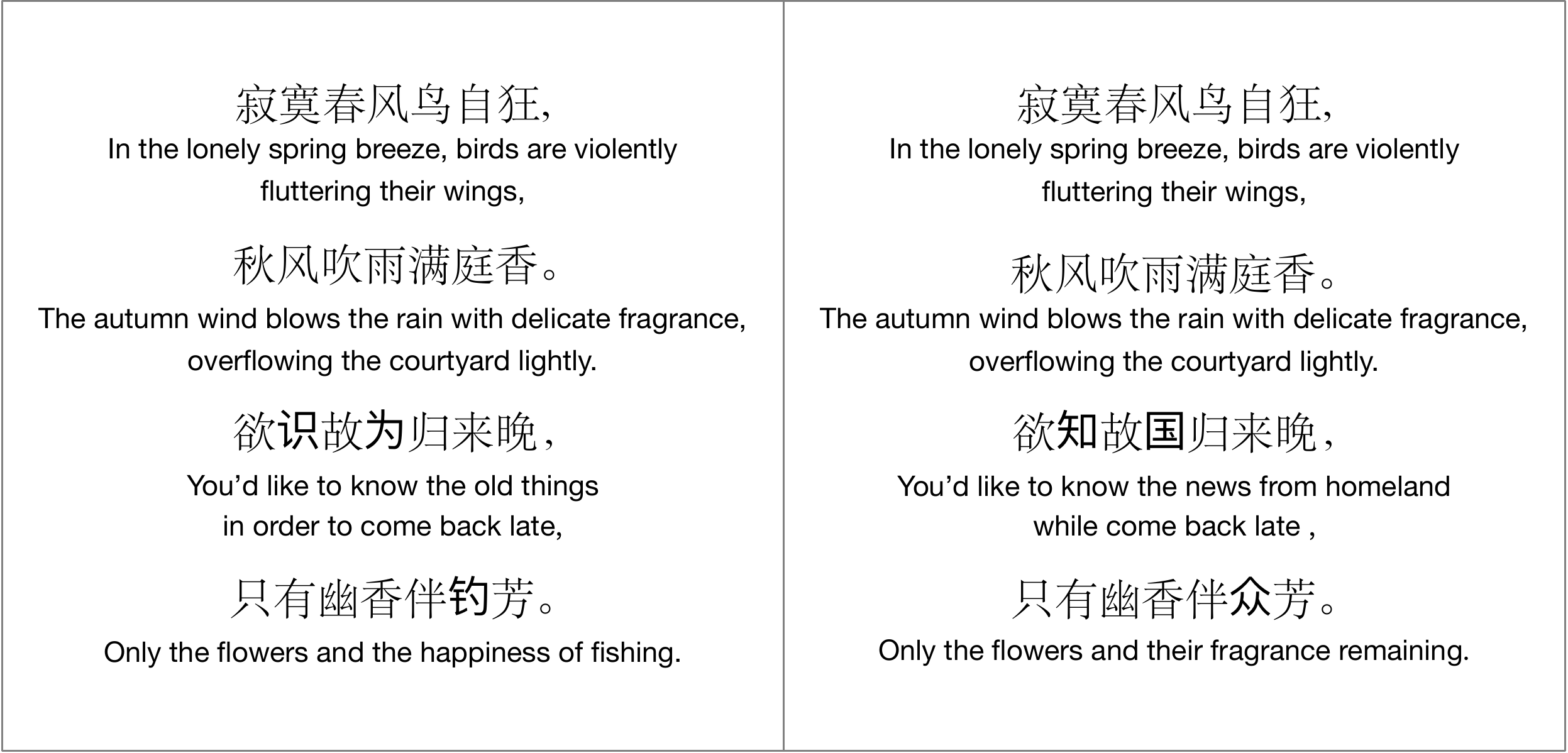}	
      \caption{The example of poem draft and the polished poem. The left poem is the poem draft generated by \textit{B\&T} and the right poem is the poem iteratively polished by \textbf{QA-MLM}. The iterative polishing is automatically terminated after 3 polish steps.}
      \label{fig:polish}
 \end{figure*}
\subsection{Results and Discussions}
The BLEU scores are compared in Table \ref{tab:bleu}.
The compared models are shown in the first column of the table, where the suffix \textbf{-P} indicates that the poems generated by previous models have been polished by \textbf{QA-MLM}.
The keywords and poem lines are extracted from test dataset with one thousand poems.
In general, the BLEU scores of \textbf{CVAE-D} are higher than \textbf{AS2S} but lower than \textbf{B\&T}, which partially reflects the generation performance of these models.
The polishing process improves the BLEU scores for \textbf{AS2S} model while has no obvious improvement or even a bit hurt of BLEU scores for \textbf{CVAE-D} and \textbf{B\&T}.
This is probably due to the creativity and diversity properties in poem generation.
The quality of generated poem is not proportional to the BLEU score when the BLEU score is comparative high.
The BLEU score should be referred conservative during the quality evaluation for poetry generation.

The other automatic evaluation results and human score results can be found in Table \ref{tab:res}.
These evaluation results are also based on one thousand poems from test dataset.
The keywords are extracted from these poems and utilized to generate poem drafts by \textbf{AS2S}, \textbf{CVAE-D} and \textbf{B\&T}.
All the poem drafts generated by this three encoder-decoder structure models have been further polished by our proposed \textbf{QA-MLM} model.
We can find that the polish procedure improves the scores of \textbf{Similarity} criteria on all the compared encoder-decoder structure models,
which demonstrates the effectiveness of our proposed \textbf{QA-MLM} for the improvement of sematic coherence and theme consistency.
The tone level and rhythm for \textbf{AS2S} and \textbf{CVAE-D} models seem randomly.
By contrast, the accuracy of tone level and rhythm for \textbf{B\&T} are significantly higher than the other two baselines, which demonstrates the embedding-based method is effective to control the tone level and rhythm.
The polishing framework on the \textbf{B\&T} hurts a bit accuracy of tone level and rhythm, which sacrifices the tone level and rhythm constraints to obtain better semantic meaning and coherence.

The human evaluations are consistent with the automatic evaluation metrics.
As for the poem draft generators, the \textbf{CVAE-D} outperforms the \textbf{AS2S} on all the quality aspects evaluated by our annotators, which is consistent with the results from \cite{li2018generating}.
However, the poem generation performance of \textbf{B\&T} is better than \textbf{CVAE-D}, which is probability due to the powerful text representation ability of BERT or transformer.

Above all, our proposed \textbf{QA-MLM} can further improve the qualities (\emph{consistency}, \emph{fluency}, \emph{meaningfulness} and \emph{poeticness} ) of poems generated by all the three aforementioned encoder-decoder structure models,
which demonstrates that the quality prediction task is effective to locate the bad characters and the masked language model task is also effective to obtain better predictions when referring to the global context information.
Therefore, the unidirectional sequential decoding deficiency of the encoder-decoder structure can be largely saved by our proposed \textbf{QA-MLM} for poetry refinement.

Although the improvements brought by the \textbf{QA-MLM} in automatic evaluation metrics seem trivial,
 the improvements are significant in the human evaluation results.
It is easy to understand that the polishing process only update a small part of characters, and the improvements will be averaged on  all the characters for automatic evaluation metrics.
However, human can understand and notice the significant difference of the changed characters to the whole poem context.
Even one character changed would lead to a big improvement for poetry quality.
We can notice the difference from the example in Figure \ref{fig:polish}.
\section{Conclusion}
In this paper, we present an iterative polishing framework for Chinese poetry generation by imitating the real poem writing process. Following the famous encoder-decoder paradigm, a pre-trained BERT encoder and a transformer decoder are combined to generate poem drafts. Then, poem polishing is accomplished by a multifunctional \textbf{QA-MLM}, which can improve poem quality in terms of semantics, syntactics and literary. Based on the multi-task learning, the trained \textbf{QA-MLM} is able to aware of the poem quality and locate improper characters.
 Besides, the \textbf{QA-MLM} is capable of predicting better ones to replace the improper characters by synthesizing the all-round poem context information.
 Moreover, the \textbf{QA-MLM} will automatically terminate the iterative polishing process when the polished draft is classified as qualified.

Both automatic evaluation and human scores demonstrate that our proposed approach is effective in Chinese poetry generation. Our model can automatically modify preliminary poems to elegant ones while keeping their original intents.
 Even though our proposed \textbf{QA-MLM} polishing framework is concentrated on Chinese poetry generation in this work, this new text refinement approach can be extended to other natural language generation areas.
\section{ Acknowledgments}
We thank Haoshen Fan, Weijing Huang, Mingkuo Ji and Shaopeng Ma for helpful discussions.

\bibliographystyle{aaai}
\bibliography{AAAI-DengL.1603}

\begin{thebibliography}{}

\bibitem[\protect\citeauthoryear{Bengio \bgroup et al\mbox.\egroup
  }{2003}]{bengio2003neural}
Bengio, Y.; Ducharme, R.; Vincent, P.; and Jauvin, C.
\newblock 2003.
\newblock A neural probabilistic language model.
\newblock {\em Journal of machine learning research} 3(Feb):1137--1155.

\bibitem[\protect\citeauthoryear{Devlin \bgroup et al\mbox.\egroup
  }{2018}]{devlin2018bert}
Devlin, J.; Chang, M.-W.; Lee, K.; and Toutanova, K.
\newblock 2018.
\newblock Bert: Pre-training of deep bidirectional transformers for language
  understanding.
\newblock {\em arXiv preprint arXiv:1810.04805}.

\bibitem[\protect\citeauthoryear{Gerv{\'a}s}{2001}]{gervas2001expert}
Gerv{\'a}s, P.
\newblock 2001.
\newblock An expert system for the composition of formal spanish poetry.
\newblock {\em Knowledge-Based Systems} 14(3-4):181--188.

\bibitem[\protect\citeauthoryear{Ghazvininejad \bgroup et al\mbox.\egroup
  }{2016}]{ghazvininejad2016generating}
Ghazvininejad, M.; Shi, X.; Choi, Y.; and Knight, K.
\newblock 2016.
\newblock Generating topical poetry.
\newblock In {\em Proceedings of the 2016 Conference on Empirical Methods in
  Natural Language Processing},  1183--1191.

\bibitem[\protect\citeauthoryear{Goldberg}{2017}]{goldberg2017neural}
Goldberg, Y.
\newblock 2017.
\newblock Neural network methods for natural language processing.
\newblock {\em Synthesis Lectures on Human Language Technologies} 10(1):1--309.

\bibitem[\protect\citeauthoryear{He, Zhou, and Jiang}{2012}]{he2012generating}
He, J.; Zhou, M.; and Jiang, L.
\newblock 2012.
\newblock Generating chinese classical poems with statistical machine
  translation models.
\newblock In {\em AAAI}.

\bibitem[\protect\citeauthoryear{Li \bgroup et al\mbox.\egroup
  }{2018}]{li2018generating}
Li, J.; Song, Y.; Zhang, H.; Chen, D.; Shi, S.; Zhao, D.; and Yan, R.
\newblock 2018.
\newblock Generating classical chinese poems via conditional variational
  autoencoder and adversarial training.
\newblock In {\em Proceedings of the 2018 Conference on Empirical Methods in
  Natural Language Processing},  3890--3900.

\bibitem[\protect\citeauthoryear{Manurung}{2004}]{manurung2004evolutionary}
Manurung, H.
\newblock 2004.
\newblock An evolutionary algorithm approach to poetry generation.

\bibitem[\protect\citeauthoryear{Mihalcea and
  Tarau}{2004}]{mihalcea2004textrank}
Mihalcea, R., and Tarau, P.
\newblock 2004.
\newblock Textrank: Bringing order into text.
\newblock In {\em Proceedings of the 2004 conference on empirical methods in
  natural language processing},  404--411.

\bibitem[\protect\citeauthoryear{Papineni \bgroup et al\mbox.\egroup
  }{2002}]{Papineni2002BLEU}
Papineni, K.; Roukos, S.; Ward, T.; and Zhu, W.~J.
\newblock 2002.
\newblock Bleu: a method for automatic evaluation of machine translation.
\newblock In {\em Proc Meeting of the Association for Computational
  Linguistics}.

\bibitem[\protect\citeauthoryear{Sun}{2012}]{sun2012jieba}
Sun, J.
\newblock 2012.
\newblock ‘jieba’chinese word segmentation tool.

\bibitem[\protect\citeauthoryear{Wang \bgroup et al\mbox.\egroup
  }{2016}]{wang2016chineseB}
Wang, Z.; He, W.; Wu, H.; Wu, H.; Li, W.; Wang, H.; and Chen, E.
\newblock 2016.
\newblock Chinese poetry generation with planning based neural network.
\newblock {\em arXiv preprint arXiv:1610.09889}.

\bibitem[\protect\citeauthoryear{Wang, Luo, and Wang}{2016}]{wang2016can}
Wang, Q.; Luo, T.; and Wang, D.
\newblock 2016.
\newblock Can machine generate traditional chinese poetry? a feigenbaum test.
\newblock In {\em International Conference on Brain Inspired Cognitive
  Systems},  34--46.
\newblock Springer.

\bibitem[\protect\citeauthoryear{Wang}{2002}]{wang2002summary}
Wang, L.
\newblock 2002.
\newblock A summary of rhyming constraints of chinese poems.

\bibitem[\protect\citeauthoryear{Wieting \bgroup et al\mbox.\egroup
  }{2015}]{wieting2015towards}
Wieting, J.; Bansal, M.; Gimpel, K.; and Livescu, K.
\newblock 2015.
\newblock Towards universal paraphrastic sentence embeddings.
\newblock {\em arXiv preprint arXiv:1511.08198}.

\bibitem[\protect\citeauthoryear{Wu, Tosa, and Nakatsu}{2009}]{wu2009}
Wu, X.; Tosa, N.; and Nakatsu, R.
\newblock 2009.
\newblock New hitch haiku: An interactive renku poem composition supporting
  tool applied for sightseeing navigation system.
\newblock In Natkin, S., and Dupire, J., eds., {\em Entertainment Computing --
  ICEC 2009},  191--196.
\newblock Berlin, Heidelberg: Springer Berlin Heidelberg.

\bibitem[\protect\citeauthoryear{Xia \bgroup et al\mbox.\egroup
  }{2017}]{xia2017deliberation}
Xia, Y.; Tian, F.; Wu, L.; Lin, J.; Qin, T.; Yu, N.; and Liu, T.-Y.
\newblock 2017.
\newblock Deliberation networks: Sequence generation beyond one-pass decoding.
\newblock In {\em Advances in Neural Information Processing Systems},
  1784--1794.

\bibitem[\protect\citeauthoryear{Yan \bgroup et al\mbox.\egroup
  }{2013}]{yan2013poet}
Yan, R.; Jiang, H.; Lapata, M.; Lin, S.-D.; Lv, X.; and Li, X.
\newblock 2013.
\newblock i, poet: Automatic chinese poetry composition through a generative
  summarization framework under constrained optimization.
\newblock In {\em IJCAI},  2197--2203.

\bibitem[\protect\citeauthoryear{Yan}{2016}]{yan2016poet}
Yan, R.
\newblock 2016.
\newblock i, poet: Automatic poetry composition through recurrent neural
  networks with iterative polishing schema.
\newblock In {\em IJCAI},  2238--2244.

\bibitem[\protect\citeauthoryear{Yang \bgroup et al\mbox.\egroup
  }{2017}]{yang2017generating}
Yang, X.; Lin, X.; Suo, S.; and Li, M.
\newblock 2017.
\newblock Generating thematic chinese poetry with conditional variational
  autoencoder.
\newblock {\em CoRR}.

\bibitem[\protect\citeauthoryear{Yi \bgroup et al\mbox.\egroup
  }{2018}]{yi2018chineseA}
Yi, X.; Sun, M.; Li, R.; and Yang, Z.
\newblock 2018.
\newblock Chinese poetry generation with a working memory model.
\newblock {\em arXiv preprint arXiv:1809.04306}.

\bibitem[\protect\citeauthoryear{Yi, Li, and Sun}{2017}]{yi2017generating}
Yi, X.; Li, R.; and Sun, M.
\newblock 2017.
\newblock Generating chinese classical poems with rnn encoder-decoder.
\newblock In {\em Chinese Computational Linguistics and Natural Language
  Processing Based on Naturally Annotated Big Data}. Springer.
\newblock  211--223.

\bibitem[\protect\citeauthoryear{Yi, Li, and Sun}{2018}]{yi2018chineseB}
Yi, X.; Li, R.; and Sun, M.
\newblock 2018.
\newblock Chinese poetry generation with a salient-clue mechanism.
\newblock {\em arXiv preprint arXiv:1809.04313}.

\bibitem[\protect\citeauthoryear{Zhang and Lapata}{2014}]{zhang2014chinese}
Zhang, X., and Lapata, M.
\newblock 2014.
\newblock Chinese poetry generation with recurrent neural networks.
\newblock In {\em Proceedings of the 2014 Conference on Empirical Methods in
  Natural Language Processing (EMNLP)},  670--680.

\bibitem[\protect\citeauthoryear{Zhang \bgroup et al\mbox.\egroup
  }{2017}]{zhang2017flexible}
Zhang, J.; Feng, Y.; Wang, D.; Wang, Y.; Abel, A.; Zhang, S.; and Zhang, A.
\newblock 2017.
\newblock Flexible and creative chinese poetry generation using neural memory.
\newblock {\em arXiv preprint arXiv:1705.03773}.

\bibitem[\protect\citeauthoryear{Zhang \bgroup et al\mbox.\egroup
  }{2019}]{zhang2019pretraining}
Zhang, H.; Gong, Y.; Yan, Y.; Duan, N.; Xu, J.; Wang, J.; Gong, M.; and Zhou,
  M.
\newblock 2019.
\newblock Pretraining-based natural language generation for text summarization.
\newblock {\em arXiv preprint arXiv:1902.09243}.

\bibitem[\protect\citeauthoryear{Zhou, You, and Ding}{2010}]{zhou2010genetic}
Zhou, C.-L.; You, W.; and Ding, X.
\newblock 2010.
\newblock Genetic algorithm and its implementation of automatic generation of
  chinese songci.
\newblock {\em Journal of Software} 21(3):427--437.

\end{thebibliography}

\end{document}